\documentclass[10pt,twocolumn,letterpaper]{article}

\usepackage{wacv}              

\usepackage{graphicx}
\usepackage{amsmath}
\usepackage{amssymb}
\usepackage{booktabs}
\usepackage{xcolor}
\usepackage{array}
\usepackage{multirow}
\usepackage[accsupp]{axessibility} 

\usepackage{physics}
\usepackage{amsmath}
\usepackage{tikz}
\usepackage{mathdots}
\usepackage{yhmath}
\usepackage{cancel}
\usepackage{color}
\usepackage{siunitx}
\usepackage{array}
\usepackage{amssymb}
\usepackage{gensymb}
\usepackage{tabularx}
\usepackage{extarrows}
\usepackage{booktabs}
\usetikzlibrary{fadings}
\usetikzlibrary{patterns}
\usetikzlibrary{shadows.blur}
\usetikzlibrary{shapes}

\usepackage[pagebackref,breaklinks,colorlinks]{hyperref}

\usepackage[capitalize]{cleveref}
\crefname{section}{Sec.}{Secs.}
\Crefname{section}{Section}{Sections}
\Crefname{table}{Table}{Tables}
\crefname{table}{Tab.}{Tabs.}

\def\wacvPaperID{515} 
\def\confName{WACV}
\def\confYear{2025}

\graphicspath{{Figures/}}

\newif\ifdraft
\draftfalse

\definecolor{orange}{rgb}{1,0.5,0}
\definecolor{pink}{rgb}{0.98, 0.38, 0.5}
\definecolor{darkgreen}{rgb}{0.055, 0.490, 0.016} 

\ifdraft
 \usepackage[normalem]{ulem}
 \newcommand{\RS}[1]{{\color{red}{\bf RS: #1}}}
 
 \newcommand{\PMN}[1]{{\color{orange}{\bf PMN: #1}}}
 
 \newcommand{\JGT}[1]{{\color{blue} JGT: #1}}
 
\else
 \newcommand{\sout}[1]{}
 \newcommand{\RS}[1]{{\color{red}{}}}
 
 \newcommand{\PMN}[1]{{\color{red}{}}}
 
 \newcommand{\JGT}[1]{{\color{blue}{}}}
 
\fi

\newcommand{\rreal}{\mathbb{R}}

\DeclareMathOperator\softmax{softmax}
\DeclareMathOperator\linear{Linear}

\newcommand{\comment}[1]{}

\newcolumntype{P}[1]{>{\centering\arraybackslash}p{#1}}

\begin{document}

\title{SAM-DA: Decoder Adapter \\for Efficient Medical Domain Adaptation}

\author{Javier Gamazo Tejero$^1$, Moritz Schmid$^1$, Pablo Márquez Neila$^1$\\ Martin S. Zinkernagel$^2$,
Sebastian Wolf$^2$, Raphael Sznitman$^1$\\
{\small     $^1$University of Bern, $^2$Inselspital Bern, Switzerland}\\
{\tt\small \{javier.gamazo-tejero, moritz.schmid, raphael.sznitman, pablo.marquez\}@unibe.ch}\\
{\tt\small \{martin.zinkernagel, sebastian.wolf\}@insel.ch}
}
\maketitle

\begin{abstract}
This paper addresses the domain adaptation challenge for semantic segmentation in medical imaging. Despite the impressive performance of recent foundational segmentation models like SAM on natural images, they struggle with medical domain images. Beyond this, recent approaches that perform end-to-end fine-tuning of models are simply not computationally tractable. To address this, we propose a novel SAM adapter approach that minimizes the number of trainable parameters while achieving comparable performances to full fine-tuning. The proposed SAM adapter is strategically placed in the mask decoder, offering excellent and broad generalization capabilities and improved segmentation across both fully supervised and test-time domain adaptation tasks. Extensive validation on four datasets showcases the adapter's efficacy, outperforming existing methods while training less than 1\% of SAM's total parameters.
\end{abstract}

\section{Introduction}
\label{sec:introduction}

Domain adaptation for semantic segmentation in medical imaging is vital to ensure that models perform effectively across different domains (\eg, different medical centers or different scanning protocols). This is critical for the practical deployment of these models in real-world medical settings where training data is often collected over a limited number of sites or settings but needs to generalize broadly. In the context of medical imaging, two domain adaptation settings are particularly interesting: (1) {\it fully supervised}, where a general model is fine-tuned with source domain images and annotations in the hope it generalizes to a target domain, and (2) {\it test-time adaptation} whereby a single target domain image is available to fine-tune a general model. While the former case needs to generalize the model to other domains via fine-tuning, the latter adapts the general model to a specific test time case. In this paper, we consider both cases. 

\begin{figure}[t]
\centering
\includegraphics[width=\columnwidth]{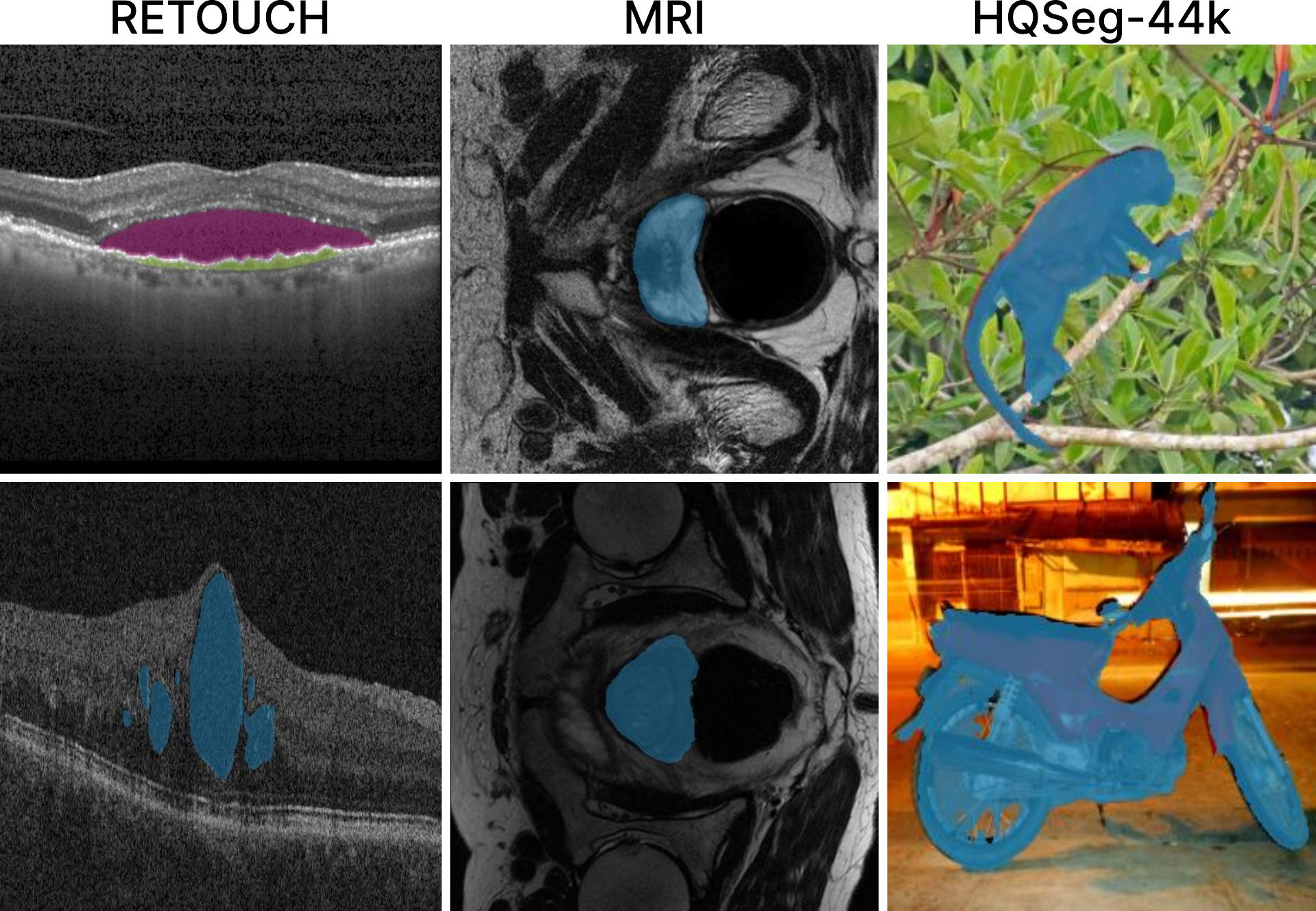}
\caption{Predictions of the proposed method on three of the four studied datasets: Retouch~\cite{retouch}, MRI~\cite{mri_dataset}, and HQSeg-44k~\cite{ke2024segment}. For the medical datasets, we show the training domain on the top row and a different domain on the bottom (Specralis and Cirrus for Retouch, BMC and UCL for MRI). For HQSeg-44k, both images come from HRSOD~\cite{hrsod_zeng2019towards}.
}
\label{fig:datasets}
\end{figure}

At the same time, recent foundational segmentation models, such as Segment Anything Model (SAM)~\cite{sam} or Segment-Everything-Everywhere Model (SEEM)~\cite{zou2024segment}, accept prompts of different input types to yield impressive zero-shot segmentation performance. Despite their capabilities on natural images, they fail to demonstrate similar prowess with medical domain images~\cite{zhang2023comprehensive}. Medical images present unique challenges due to their inherent characteristics: typically lower contrast compared to natural images, the potential for the object of interest to blend with the background in terms of color, and segmentation masks with potentially irregular morphologies. Other medical modalities, such as MRI or CT, are 3D instead of 2D, and adding such information to existing architectures is not straightforward. Researchers have addressed these limitations by fully fine-tuning the model end-to-end specifically for medical data~\cite{MedSAM,zhang2023customized}. However, this approach incurs significant computational and memory costs, and the actual benefit from training end-to-end is in question, as pre-trained models on natural images often transfer to medical images~\cite{raghu2019transfusion,zamir2018taskonomy}. Moreover, while this approach may offer a temporary workaround, the ongoing expansion of models and the scarcity of medical data present persistent challenges. These factors suggest that end-to-end training of future models may only be feasible for entities with ample resources.

In response to this challenge, we have witnessed the emergence of Parameter Efficient Fine-Tuning (PEFT) methods. These techniques aim to minimize the number of trainable parameters while achieving comparable performance to full fine-tuning~\cite{xu2023parameter,hu2022lora,pfeiffer2020adapterfusion,mahabadi2021parameter,lin2020exploring}. Originally developed within the domain of Natural Language Processing (NLP), these techniques are now experiencing rapid evolution, driven by the emergence of increasingly large language models. This trend is fueled by the impractical training times associated with these larger models, necessitating more efficient fine-tuning strategies\cite{li2021prefix,liu2023gpt,he2021towards}. 

Another aspect of medical data lies in its variety of domains stemming from the vast diversity of acquisition protocols and devices. It remains uncertain whether a model trained on specific device settings will exhibit comparable performance on a different domain, even when confronted with the same image modality. Therefore, developing a model capable of generalizing over unseen domains is invaluable, not only for practical reasons but also because retraining a model on a different domain would necessitate further certification of the new model. In this line, recent efforts have been devoted to Test-Time Domain Adaptation for semantic segmentation~\cite{janouskova2023single,wang2023dynamically}, a scenario in which a trained model is fine-tuned for a single sample. 

We propose a novel SAM adapter that offers excellent and broad generalization capabilities due to its strategic placement in the mask decoder while simultaneously yielding improved segmentation across both fully supervised and test-time domain adaptation tasks. Our adapter is simple in nature and leverages the pre-trained model that inherently contains domain knowledge. Consequently, neither the image encoder nor the mask decoder require significant parameter updates during the adaptation phase. By making this design choice, we significantly decrease the number of trainable parameters compared to existing methods, making it highly efficient and easy to train. We extensively validate our approach across three medical datasets and one natural image dataset. In addition, we provide comprehensive ablation studies that explore the impact of our design choices. Our results demonstrate that the SAM Decoder Adapter (SAM-DA) outperforms general methods such as LoRA~\cite{hu2022lora} and HQ-SAM~\cite{ke2024segment}, and also medical-specific methods (\eg~ Med-SA~\cite{wu2023medical}) on both fully supervised segmentation and test-time domain adaptation. Particularly noteworthy is that this superior performance is achieved by training less than $1\%$ of the total SAM parameters.

Our contributions can be summarized as follows:
\begin{itemize}
    \item We propose an adapter for SAM based on previous LLM literature. We position it in the decoder, significantly reducing parameters and training time.
    \item We validate our setting on two tasks: domain generalization and test-time domain adaptation. 
    \item Our experiments show that the proposed adapter achieves better generalization than other encoder-focused PEFT methods. 
\end{itemize}

\section{Related work}
\label{sec:related}
\subsection{Parameter-Efficient Fine-Tuning} 
Parameter Efficient Fine-Tuning (PEFT) methods have emerged as a response to the increasing model size. These techniques aim to minimize the number of trainable parameters while achieving comparable performance to full fine-tuning. By training only a subset of parameters, PEFT methods aim to retain the knowledge of the base model when trained on a secondary task, thereby mitigating issues like catastrophic forgetting and overfitting, especially when dealing with smaller target datasets. The landscape of PEFT methodologies is vast~\cite{xu2023parameter}, with LoRA~\cite{hu2022lora} standing out as a resilient method over time. LoRA incorporates two trainable low-rank matrices for weight updates. Specifically, it employs a down-projection matrix and an up-projection matrix alongside the query, key, and value matrices within the attention layer of the transformer.

Others have envisioned adaptation methods that function as plugins for large models. These emerged along with large-scale models in the Natural Language Processing literature~\cite{houlsby2019parameter} and have subsequently spread to other fields of Deep Learning. The fundamental concept underlying these methods is to insert a module with few parameters into the base model and solely update those while maintaining the pre-trained model frozen. Initially pioneered by the Adapter framework~\cite{houlsby2019parameter}, this approach inserts such modules sequentially after the self-attention layer in all transformer layers. Since then, various other methodologies have emerged, with adaptations in the position of the adapter~\cite{pfeiffer2020adapterfusion,mahabadi2021parameter,lin2020exploring}. This evolution has also transpired in semantic segmentation, where large models are beginning to establish their significance~\cite{chen2022vision,chen2023sam}. The medical domain presents formidable challenges and has witnessed recent advancements, exemplified by works like Medical SAM Adapter~\cite{wu2023medical}. Here, Wu et al. develop three methods in increasing complexity: firstly, they incorporate a traditional adapter into the image encoder; secondly, they extend adapter functionality to accommodate 3D images by duplicating certain layers; finally, they propose a prompt-conditioned adaptation approach. More recently, \cite{ke2024segment} proposes a method to improve the quality of SAM segmentation masks via a learnable High-Quality Output Token injected into SAM's decoder that receives features from the ViT image encoder.

Lastly, certain methods leverage the tokenization process of LLMs and implement PEFT with prompt tuning~\cite{lester2021power,zhou2022learning,llama_adapter,xing2023dual}. One such method, LLaMA-Adapter~\cite{llama_adapter}, was explicitly developed to fine-tune LLaMA~\cite{llama} into an instruction-following model. Specifically, in the higher transformer layers of LLaMA, they append a set of trainable, zero-initialized adaptation prompts as a prefix to the input instruction tokens. 

\subsection{Domain Adaptation for Semantic Segmentation}
Domain adaptation has a rich history owing to its practical utility in reducing annotation requirements in unseen domains. Techniques like Unsupervised Domain Adaptation aim to do so without target labels~\cite{ghamsarian2023domain,hoyer2023mic,chen2023pipa}, Source Free Domain Adaptation restricts the task even more by removing access to the source domain annotated data~\cite{zhang2023improving,kundu2020universal,xia2021adaptive}, and Test-Time Domain Adaptation (TTDA) focuses on a setting in which data is received online and the model has to adapt per sample~\cite{janouskova2023single,wang2023dynamically}. Large models aim to mitigate the need for domain adaptation by training on extensive datasets, allowing them to generalize to unseen domains from others that are present in the dataset~\cite{oquab2023dinov2}. To our knowledge, the effectiveness of these approaches when applied to SAM has not been extensively investigated, with only a few studies such as \cite{zhang2023improving} exploring this area. In this work, the authors propose a weakly supervised self-training architecture to enhance the robustness and computation efficiency of the adaptation.

\section{Method}
\label{sec:method}
We propose a simple, lightweight adapter for the Segment Anything Model (SAM)~\cite{sam} that draws inspiration from adaptation techniques in the NLP literature. 


\subsection{Segment Anything Model}
\label{sec:sam}
SAM consists of three primary components: an image encoder, a prompt encoder, and a mask decoder. The image encoder is a standard MAE pre-trained Vision Transformer (ViT)~\cite{dosovitskiy2021an} that transforms the input image into an embedding space. The prompt encoder takes either sparse (points, boxes) or dense (masks) annotation formats and produces encoded prompts. Both the image embedding and the encoded prompts are fed to the mask decoder, which consists of a transformer block with a mask prediction head. The transformer block applies two layers of two-way cross-attention operations acting on the image and the prompt embeddings. The result of the transformer is upsampled with an MLP and then fed to a linear classifier, which predicts the final segmentation mask.

\subsection{SAM Decoder Adapter}
\label{sec:adapter}

The LLaMA-Adapter~\cite{llama_adapter} is an adaptation method originally introduced to finetune pre-trained LLMs such as LLaMA~\cite{llama}. In the context of language generation, this adapter introduces a set of learnable adaptation prompts at the higher layers of the LLaMA transformer, and prepends them to the word tokens before the attention operations. 

\begin{figure*}[t]
\centering
\includegraphics[width=\textwidth]{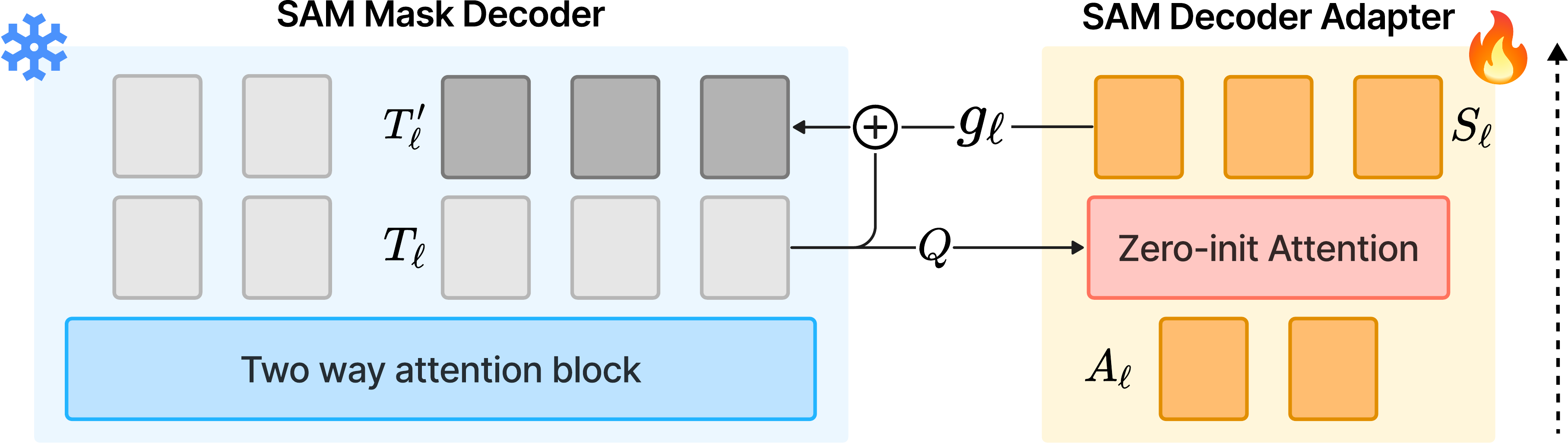}
\caption{Illustration of the proposed adaptation for SAM Decoder in layer~$\ell$. In each layer, the adaptation embeddings $A_\ell$ are fed along with the dense embeddings $T_\ell$ to the trainable zero-initialized attention module, where the dense embeddings $T_\ell$ act as queries and the adaption $A_\ell$, as keys and values. Then, the resulting tokens $S_\ell$ are projected back to the model dimension with a linear layer (omitted in the figure) and finally combined with the decoder embeddings via a trainable gating parameter $g_\ell$ and a linear MLP, resulting in $T'_\ell$, which substitutes the previous dense embeddings. A detailed neural circuit diagram~\cite{abbott2024neural} can be found in the supplementary material.
}
\label{fig:method}
\end{figure*}

Our approach brings the idea of prompt-based adaptation from NLP to SAM. We introduce a new learnable adaptation prompt~$A_\ell\in\rreal^{N\times{}D}$ at each layer~$\ell$ of the mask decoder's transformer. The adaptation prompts are used to compute correction factors that modify the embeddings of the transformer without retraining its parameters. Formally, let~$T_\ell\in\rreal^{M\times D}$ be the embeddings obtained as the output of the cross-attention operation at layer~$\ell$.
We feed $A_\ell$ and~$T_\ell$ to an additional attention block, where the embeddings $T_\ell$ act as queries and the adapter weights~$A_\ell$ act as keys and values,
\begin{eqnarray}
    Q_\ell & = & \linear^q_\ell(T_{\ell}) \in \rreal^{M\times D_k}, \\
    K_\ell & = & \linear^k_\ell(A_{\ell}) \in \rreal^{N\times D_k}, \\
    V_\ell & = & \linear^v_\ell(A_{\ell}) \in \rreal^{N\times D_v}.
\end{eqnarray}
The attention scores are calculated as usual,
\begin{equation}
    S_\ell = \softmax\left(\dfrac{Q_\ell{}K_\ell^T}{\sqrt{D_v}}\right)V_\ell \in \rreal^{M\times D_v},
\end{equation}
and projected back to the model dimension~$D$ with a linear layer, $S'_\ell=\linear^o_\ell(S_\ell)$. The result~$S'_\ell$ serves as the correction factor of the original embeddings,
\begin{equation}
    T'_\ell = \linear^t_\ell(T_\ell + g_\ell \cdot S'_\ell),
\end{equation}
where the learnable gating factor~$g_\ell\in\rreal$ is initialized to~0 to ensure no disruption during the early stages of adaptation learning. Therefore, $T'_\ell$ substitutes the previous dense embeddings $T_\ell$ as the input for the two-way attention block in the next layer $\ell + 1$. The entire procedure is summarized in Fig.~\ref{fig:method}.

We note that this approach could potentially be implemented within the model's encoder. However, we deliberately decided not to pursue this route for two main reasons: (1) the image representation generated by the encoder is already high-quality due to a pre-trained model on similar data, and (2) any modifications to the encoder's parameters may necessitate retraining of the mask decoder as well. Given our objective of reducing the number of parameters, we confine the implementation solely to the decoder. In Sec.~\ref{sec:ablation}, we show experimentally how the location of the adapter affects the model's performance.

\section{Experimental setup}
\label{sec:experiments}
The following section details our experimental setup and compares our approach to several baselines. We apply our method to two scenarios: fully-supervised semantic segmentation and test-time domain adaptation (TTDA) for semantic segmentation on four datasets.

\subsection{Datasets}
We validate our approach on one natural image dataset and three medical cross-device/site datasets:
\begin{itemize}
    \item{\textbf{Retouch}~\cite{retouch}}: Retinal OCT volumes from three devices - Cirrus, Spectralis, and Topcon devices - including segmentation masks for three biological markers: IRF, SRF, and PED. We use ``Spectralis'' dataset as the source and ``Cirrus'' dataset as the target domain. 
    
    
    \item{\textbf{MRI}~\cite{liu2020shape,mri_dataset}}: Prostate MRI scans from various sites with different devices showing prostate capsule segmentation at various cancer stages. We extract subsets from Boston Medical Center (BMC) and University College London (UCL). ``BMC'' serves as the source domain, while ``UCL'' is the target.
    
    
    \item{\textbf{WMH}~\cite{wmh}}: Brain MRI scans from three sites: NUHS Singapore, UMC Utrecht, and VU Amsterdam, with White Matter Hyperintensities and other pathologies segmented. We extract FLAIR axial slices containing segmentation. We sample volumes from ``UMC Utrecht'' and ``NUHS Singapore'' as the source and target domain, respectively. 
    

    \item{\textbf{HQSeg-44K}~\cite{ke2024segment}} contains a collection of datasets for training and testing. For training, data is taken from DIS~\cite{dis_qin2022highly} (train set), ThinObject-5K~\cite{coift_liew2021deep} (train set), FSS-1000~\cite{fss_li2020fss}, ECSSD~\cite{eccsd_shi2015hierarchical}, MSRA10K~\cite{msra_cheng2014global} and DUT-OMRON~\cite{dutomron_yang2013saliency}. The models are tested on a collection of four datasets: DIS~\cite{dis_qin2022highly} (validation set), ThinObject-5K~\cite{coift_liew2021deep} (test set), COIFT~\cite{coift_liew2021deep}, and HRSOD~\cite{hrsod_zeng2019towards}. These datasets contain fine-grained mask labels of natural images and together they add up to more than 1,000 semantic classes. 
    
\end{itemize}
In all cases, we use the source domain as the only domain for the fully supervised experiments. Since HQSeg-44K contains natural images with no clear domain shift, we do not report results for generalization and domain adaptation on this dataset. We report all the results on separate test sets, following hyperparameter fine-tuning on a validation set and repeating each experiment four times. Table \ref{tab:dataset_details_samda} describes the size of the training, validation, and test splits for each dataset and domain. 

\begin{table*}[]
\centering
\begin{tabular}{@{}p{2.2cm}lP{2.2cm}P{2.2cm}P{2.2cm}@{}}
\toprule
\multicolumn{1}{l}{\textbf{Dataset}} & \textbf{Domains} & \textbf{Training} & \textbf{Validation} & \textbf{Test} \\ \midrule
\multirow{2}{*}{Retouch} & Spectralis & 1,773 & 591 & 480 \\
 & Cirrus & 3,765 & 1,255 & 1,252 \\ \midrule
\multirow{2}{*}{MRI} & BMC & 177 & 92 & 93 \\
 & UCL & 82 & 41 & 41 \\ \midrule
\multirow{2}{*}{WMH} & Utrecht & 4,260 & 1,065 & 1,255 \\
 & Singapore & 3,560 & 890 & 1,105 \\ \midrule
HQSeg-44k & - & 44,320 & \multicolumn{2}{c}{1,537} \\
 
 \bottomrule
\end{tabular}
\caption{Number of images for each dataset per split. Since HQSeg-44K is a combination of datasets, it uses the same set for validation and testing.}
\label{tab:dataset_details_samda}
\end{table*}

\subsection{Baselines}
We compare our adapter against four alternative techniques. Among these, three are prominent examples of contemporary state-of-the-art approaches: LoRA~\cite{hu2022lora}, Med-SA~\cite{wu2023medical} and HQ-SAM~\cite{ke2024segment}. The remaining methods involve completely fine-tuning the model and solely fine-tuning the decoder while keeping the encoder frozen. It is important to acknowledge that certain methods alter the image embedding as they are integrated within the encoder. To ensure equitable comparisons, we train both the adapter and the mask decoder in these cases. We use SAM with the ViT-B/16~\cite{dosovitskiy2021an} variant of the Vision Transformer as the encoder, pre-trained with MedSAM~\cite{MedSAM} weights for all our experiments on medical datasets and the official SAM pre-training weights for HQSeg-44K. All the baseline methods use the SAM architecture, and LoRA weights are placed in the encoder.



\subsection{Fully Supervised Training Experiments}
Here, we evaluate SAM Decoder Adapter through fully supervised semantic segmentation training on individual images and evaluate the models on the same training domain and a different domain. 
We use AdamW~\cite{loshchilov2017decoupled} optimizer in all cases with a loss that combines a \emph{mask prediction loss} and a \emph{IoU prediction loss}. The mask prediction loss is a linear combination of Dice loss~\cite{milletari2016dice} and Cross Entropy loss in a 0.8:0.2 ratio. The IoU prediction loss is used to train the IoU prediction head of SAM. It is computed as the MSE between the IoU prediction of SAM and the IoU of the predicted mask with the ground truth mask. As reported in~\cite{sam}, using this IoU prediction loss with a weight of 1.0 increases performance slightly. Other hyperparameters are fine-tuned with the validation set whenever possible (Retouch, MRI, and WMH). For HQSeg-44k, we use the same hyperparameters as described in HQ-SAM~\cite{ke2024segment}. Our adapter uses $N=2$~tokens and dimension~$D=512$. The dimensions of the attention module are~$D_v=D_k=256$.

\subsection{Test-Time Domain Adaptation Experiments}

Test-time domain adaptation (TTDA) refers to a scenario in which a model receives unlabeled data during inference from a dataset with a data distribution that differs from that on which it was trained. Accordingly, the model must adapt on a sample-by-sample basis, attempting to extract the maximum amount of information from each data point. In these experiments, we assume that the domain shift between the source and target distributions is due solely to differences in the acquisition device and, therefore, that the class semantics and number of classes remain unchanged. This scenario is not hypothetical in clinical practice: a model trained to detect biological markers with a specific configuration of the image acquisition device can be required to adapt to a new configuration, but the underlying detection scheme (number and semantics of classes) will remain the same.

We adopt a conventional training approach centered on entropy minimization per sample, leveraging the most confident samples~\cite{wang2021tent}. Then, we enforce proximity between the output and initial predictions through a regularization term incorporating focal loss~\cite{lin2017focal} and dice loss, following~\cite{zhang2023improving}. Finally, for Retouch, we introduce a contrastive term within the volume. Negative slices are sampled distantly from the anchor, while positive slices are nearby. We optimize the contribution of each term to the final loss for every method and dataset on the validation set, as well as the number of iterations. We zero-initialize the adapter of the Med-SA baseline, ensuring that it does not affect the model predictions before adaptation. Due to the challenge of zero-initializing HQ-SAM, we omit it from these experiments.

Note that the adapter that we propose in this paper is agnostic to the test-time domain adaptation algorithm. For this reason, we opt to use a simple training approach that will not shade the adapter's capabilities. Furthermore, comparisons are only carried out against other adapters and PEFT methods. Comparing against different test-time domain adaptation techniques would be out of the scope of the present work.

\section{Results}
\label{sec:results}

\begin{table*}[t]
\centering
\begin{tabular}{@{}lP{3cm}P{2.3cm}P{2.3cm}P{2.3cm}|P{3.2cm}@{}}
\toprule
  & Retouch - Spectralis & MRI - BMC & WMH - Utrecht & HQSeg-44K & Learnable Params (M) \\ \midrule
Fine-Tuning &  $76.0{\scriptscriptstyle \pm 0.6}$ & $85.9{\scriptscriptstyle \pm 1.5}$ & $43.5{\scriptscriptstyle \pm 0.9}$ & $76.0{\scriptscriptstyle \pm 0.3}$ & 90.60 \\ \midrule
Decoder FT  & $42.8{\scriptscriptstyle \pm 2.0}$ & $71.9{\scriptscriptstyle \pm 2.6}$ & $40.8{\scriptscriptstyle \pm 0.3}$ & $80.9{\scriptscriptstyle \pm 0.3}$ & 3.92 \\
LoRA~\cite{hu2022lora} & $74.1{\scriptscriptstyle \pm 1.1}$ & $83.7{\scriptscriptstyle \pm 1.3}$ & $43.1{\scriptscriptstyle \pm 0.5}$ & $83.1{\scriptscriptstyle \pm 0.3}$* & 4.07 \\
Med-SA~\cite{wu2023medical}  & $75.0{\scriptscriptstyle \pm 1.2}$* & $84.3{\scriptscriptstyle \pm 1.9}$* & $\mathbf{44.7}{\scriptscriptstyle \pm 0.7}$ & $\mathbf{83.8}{\scriptscriptstyle \pm 0.4}$ & 11.03 \\
HQ-SAM~\cite{ke2024segment}  & $52.4{\scriptscriptstyle \pm 2.1}$ & $76.5{\scriptscriptstyle \pm 1.5}$ & $40.9{\scriptscriptstyle \pm 0.6}$ & $79.2{\scriptscriptstyle \pm 0.2}$ & 1.07 \\ \midrule
SAM-DA  & $\mathbf{75.4}{\scriptscriptstyle \pm 0.6}$ & $\mathbf{86.2}{\scriptscriptstyle \pm 1.5}$ & $44.2{\scriptscriptstyle \pm 0.3}$* & $79.6{\scriptscriptstyle \pm 0.4}$ & 0.66 \\ \bottomrule
\end{tabular}
\caption{IoU scores for the full supervision task. Variances are computed over four trained models tested on the testing set. Bold numbers indicate the best adapter. Asterisks indicate the second best. Learnable parameters refer to the number of parameters that are trained for each method.}
\label{tab:supervised_training}
\end{table*}

\subsection{Full Supervision}
\begin{figure*}[t]
\centering
\includegraphics[width=\textwidth]{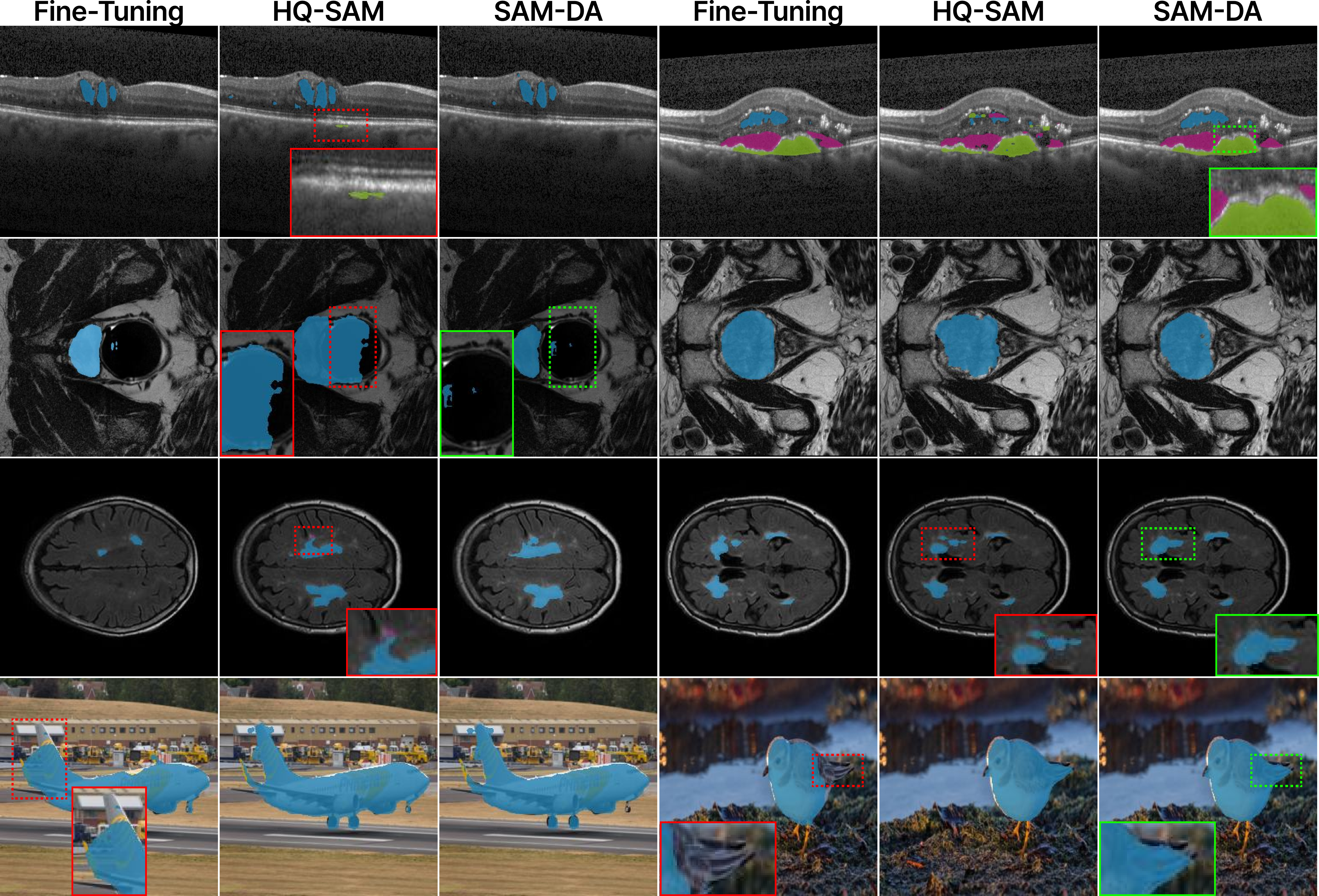}
\caption{Qualitative results on eight randomly selected in-domain test samples.}
\label{fig:full_superv}
\end{figure*}

Table~\ref{tab:supervised_training} presents a comparison of the IoU scores achieved by SAM Decoder Adapter (SAM-DA) against alternative baselines across four datasets in the fully supervised task. Our approach consistently delivers comparable or superior results to full fine-tuning despite employing only a fraction of the trainable parameters. Notably, the number of training images influences the model's final performance: the BMC domain in the MRI dataset, with the fewest samples across all datasets, showcases significant performance improvement with our adapter compared to other methods. With fewer parameters, SAM Decoder Adapter is less susceptible to overfit to the training set, thereby retaining valuable knowledge from the pre-trained weights. This effect diminishes as the dataset size increases, where the advantage of the adapter over competitive baselines is less evident. Fig.~\ref{fig:full_superv} shows qualitative results of our adapter.

Due to the substantial number of images in HQSeg-44K~\cite{ke2024segment}, this dataset can be considered quite distinct. With 44k training images, encoder-adaptation methods are expected to outperform decoder-only approaches due to their higher number of parameters. Our method achieves an IoU score of 79.6 on this dataset, falling behind LoRA and Med-SA (with an average IoU of 83.5). However, it surpasses other decoder-only adapters like HQ-SAM, which achieves an IoU of 79.2. These results are the average over the four testing sets in HQSeg-44K (see supplementary material for further figures). 

\subsection{Domain Generalization}
One of the primary strengths of SAM lies in its ability to zero-shot transfer thanks to its extensive pre-training corpus. We investigate in Table~\ref{tab:zero_shot} whether this capability is retained after the adaptation by testing the methods from Table~\ref{tab:supervised_training} on previously unseen domains within each medical dataset. SAM Decoder Adapter demonstrates statistically significant superiority in zero-shot generalization compared to other methods that primarily focus on traditional encoder adaptation, such as Med-SA and LoRA, and shows on-par performance to fine-tuning in the low-data regime. See supplementary material for qualitative results of our adapter compared to two baselines on the generalization domains.

\begin{table}[t]
\centering
\begin{tabular}{@{}lP{1.8cm}P{1.6cm}P{1.6cm}@{}}
\toprule
 & Retouch - Cirrus & MRI - UCL & WMH - Singapore \\ \midrule
Fine-Tuning & $65.4{\scriptscriptstyle \pm 5.1}$ & $80.9{\scriptscriptstyle \pm 1.1}$ & $40.1{\scriptscriptstyle \pm 0.9}$ \\ \midrule
Decoder FT & $27.0{\scriptscriptstyle \pm 0.7}$ & $70.4{\scriptscriptstyle \pm 0.8}$ & $35.8{\scriptscriptstyle \pm 0.95}$ \\
LoRA~\cite{hu2022lora} & $56.0{\scriptscriptstyle \pm 5.6}$ & $75.9{\scriptscriptstyle \pm 1.5}$ & $37.3{\scriptscriptstyle \pm 1.0}$ \\
Med-SA~\cite{wu2023medical} & $61.7{\scriptscriptstyle \pm 3.5}$ & $75.8{\scriptscriptstyle \pm 7.0}$ & $38.8{\scriptscriptstyle \pm 1.2}$ \\
HQ-SAM~\cite{ke2024segment} & $30.8{\scriptscriptstyle \pm 0.4}$ & $75.8{\scriptscriptstyle \pm 3.7}$ & $35.9{\scriptscriptstyle \pm 0.7}$ \\ \midrule
SAM-DA & $\mathbf{70.2}{\scriptscriptstyle \pm 3.1}$ & $\mathbf{80.6}{\scriptscriptstyle \pm 1.0}$ & $\mathbf{39.6}{\scriptscriptstyle \pm 0.7}$ \\ \bottomrule
\end{tabular}
\caption{Domain generalization results on an unseen domain (IoU score). Variances are computed over four trained models tested on the testing set.}
\label{tab:zero_shot}
\end{table}


\subsection{Ablation studies}
\label{sec:ablation}
We attribute the significant generalization capability of our method to its focus on adapting the decoder only, and prove it by applying it to the encoder. Unlike other approaches that prioritize encoder adaptation, necessitating decoder training, our method leverages pre-trained weights, yielding an already proficient decoder. The Vision Transformer-type architecture used in the backbone of SAM facilitates the seamless integration of the proposed adapter into the encoder with minimal adjustments compared to the configuration depicted in Fig.~\ref{fig:method}. We adopt the approach outlined in~\cite{llama_adapter} to position the adapter within the encoder, focusing on adapting only the last layers. Additionally, we utilize all image embeddings as queries for the attention module and fine-tune the decoder, following the setting from LoRA and Med-SA. ViT-B comprises 12 blocks, and we modify the last 10 blocks in our adaptation approach. This adaptation strategy allocates $17.7$M trainable parameters to the encoder adaptation, in addition to $3.9$M parameters in the decoder. 

\begin{table}
\centering
\begin{tabular}{@{}lP{1.6cm}P{1.15cm}P{1.15cm}P{1.55cm}@{}}
\toprule
 & Retouch - Spectralis & MRI - BMC & WMH - Utrecht & HQSeg-44K  \\ \midrule
Decoder & $75.4{\scriptscriptstyle \pm 0.6}$ & $\mathbf{86.2}{\scriptscriptstyle \pm 1.5}$ & $\mathbf{44.2}{\scriptscriptstyle \pm 0.3}$ & $79.6{\scriptscriptstyle \pm 0.4}$ \\
Encoder & $\mathbf{75.8}{\scriptscriptstyle \pm 1.0}$ & $84.6{\scriptscriptstyle \pm 2.2}$ & $40.7{\scriptscriptstyle \pm 0.3}$ & $\mathbf{80.8}{\scriptscriptstyle \pm 0.3}$ \\ \bottomrule 
\end{tabular}
\caption{Ablation study evaluating two adapter locations. IoU scores for full supervision. Variances are computed over four trained models tested on the testing set.}
\label{tab:encoder_ablation_trained}
\end{table}

Tables~\ref{tab:encoder_ablation_trained} and~\ref{tab:encoder_ablation_transfer} illustrate the impact of adapting the encoder layers compared to the proposed method. With respect to fully supervised performance, the results in Table~\ref{tab:encoder_ablation_trained} confirm the previous finding that adapting the decoder has diminished returns as the dataset size increases, with Retouch and HQSeg showing significantly higher performance with encoder adaptation. On the other hand, Table~\ref{tab:encoder_ablation_transfer} shows that locating the adapter in the decoder improves generalization on unseen domains, especially in the case of Retouch and WMH datasets, where we see a gain of $8.5\%$ and $13.8\%$ in IoU score, respectively. 

\begin{table}[h]
\centering
\begin{tabular}{@{}lP{2cm}P{1.6cm}P{1.8cm}@{}}
\toprule
 & Retouch - Cirrus & MRI - UCL & WMH - Singapore \\ \midrule
Decoder & $\mathbf{70.2}{\scriptscriptstyle \pm 3.1}$ & $\mathbf{80.6}{\scriptscriptstyle \pm 1.0}$ & $\mathbf{39.6}{\scriptscriptstyle \pm 0.7}$ \\
Encoder & $64.7{\scriptscriptstyle \pm 6.8}$ & $80.4{\scriptscriptstyle \pm 1.4}$ & $34.8{\scriptscriptstyle \pm 0.6}$ \\ \bottomrule
\end{tabular}
\caption{Ablation study of two adapter locations. Application of trained models to zero-shot domain generalization. The variance was obtained over four trained models tested on the testing set.}
\label{tab:encoder_ablation_transfer}
\end{table}

Tables~\ref{tab:ablation_size} and~\ref{tab:ablation_size_transfer} show the impact of the dimension of the adaptation prompt $A_\ell$ on the performance. The results do not suggest that the size of the adaptation prompt impacts the full supervision or the zero-shot generalization performance for the medical datasets significantly. For HQSeg-44K, however, the drop in performance when the size is increased is blatant. This is in line with the previous observation that decoder-only methods cannot trace encoder-decoder approaches in large datasets, and further adding parameters only promotes overfitting to the training set.

\begin{table}[]
\centering
\begin{tabular}{@{}lP{1.6cm}P{1.15cm}P{1.15cm}P{1.55cm}@{}}
\toprule
 & Retouch - Spectralis & MRI - BMC & WMH - Utrecht & HQ-Seg \\ \midrule
512 & $75.4{\scriptscriptstyle \pm 0.6}$ & $86.2{\scriptscriptstyle \pm 1.5}$ & $44.2{\scriptscriptstyle \pm 0.3}$ & $79.6{\scriptscriptstyle \pm 0.4}$ \\
1024 & $75.8{\scriptscriptstyle \pm 0.8}$ & $85.6{\scriptscriptstyle \pm 1.6}$ & $40.5{\scriptscriptstyle \pm 3.8}$ & $71.6{\scriptscriptstyle \pm 2.8}$ \\
2048 & $76.3{\scriptscriptstyle \pm 0.8}$ & $85.2{\scriptscriptstyle \pm 1.7}$ & $39.0{\scriptscriptstyle \pm 4.7}$ & $70.2{\scriptscriptstyle \pm 2.3}$ \\ \bottomrule
\end{tabular}
\caption{Ablation study for the size of the adapter embeddings. IoU scores for full supervision. Variances are computed over four trained models tested on the testing set.}
\label{tab:ablation_size}
\end{table}

\begin{table}[t]
\centering
\begin{tabular}{@{}lP{2cm}P{1.6cm}P{1.8cm}@{}}
\toprule
 & Retouch - Cirrus & MRI - UCL & WMH - Singapore \\ \midrule
512 & $70.2{\scriptscriptstyle \pm 3.1}$ & $80.6{\scriptscriptstyle \pm 1.0}$ & $39.6{\scriptscriptstyle \pm 0.7}$ \\
1024 & $69.4{\scriptscriptstyle \pm 1.3}$ & $79.8{\scriptscriptstyle \pm 1.3}$ & $38.7{\scriptscriptstyle \pm 3.1}$ \\
2048 & $69.2{\scriptscriptstyle \pm 1.2}$ & $80.2{\scriptscriptstyle \pm 1.1}$ & $35.5{\scriptscriptstyle \pm 3.4}$ \\ \bottomrule
\end{tabular}
\caption{Ablation study for the size of the adapter embeddings. Application of trained models to zero-shot domain generalization. The variance was obtained over four trained models tested on the testing set.}
\label{tab:ablation_size_transfer}
\end{table}

\begin{table}[t]
\centering
\begin{tabular}{@{}lP{1.8cm}P{1.6cm}P{1.8cm}@{}}
\toprule
 & Retouch - Cirrus & MRI - UCL & WMH - Singapore \\ \midrule
Fine-Tuning & 61.2 & 80.2 & 39.6 \\
Decoder FT & 66.9 & 80.8 & 40.4 \\
LoRA~\cite{hu2022lora} & 67.0 & 80.7 & 40.3 \\
Med-SA~\cite{wu2023medical} & 63.3 & 80.7 & 40.4 \\ \midrule
SAM-DA & \textbf{67.5} & \textbf{81.1} & \textbf{40.4} \\ \bottomrule
\end{tabular}
\caption{Test-time domain adaptation results (IoU score). Due to the challenge of zero-initializing HQ-SAM, we omit it from these experiments.}
\label{tab:ttda}
\end{table}

\subsection{Test-Time Domain Adaptation}
Due to the low number of trainable parameters, test-time domain adaptation is the ideal setting to test how much an adapter can affect the performance of a single test sample. We show in Table~\ref{tab:ttda} that our adapter performs better than other baselines in most cases. For MRI, the increase in trainable parameters already in LoRA hurts the initial model and decreases its IoU score after only five iterations. Comparing fine-tuning against other approaches in said Table, we see that the addition of parameters penalizes performance. A higher number of parameters increases the capacity of the model and, therefore, its ability to learn. However, this is a double-edged sword in test-time domain adaptation, as the network may also learn noise or characteristics of the data that are not representative. This behavior is exacerbated in unsupervised learning done via entropy minimization, where the signal is intrinsically noisy. Our method, however, shows strong results in all three cases.

\begin{figure*}[t]
\centering
\includegraphics[width=\textwidth]{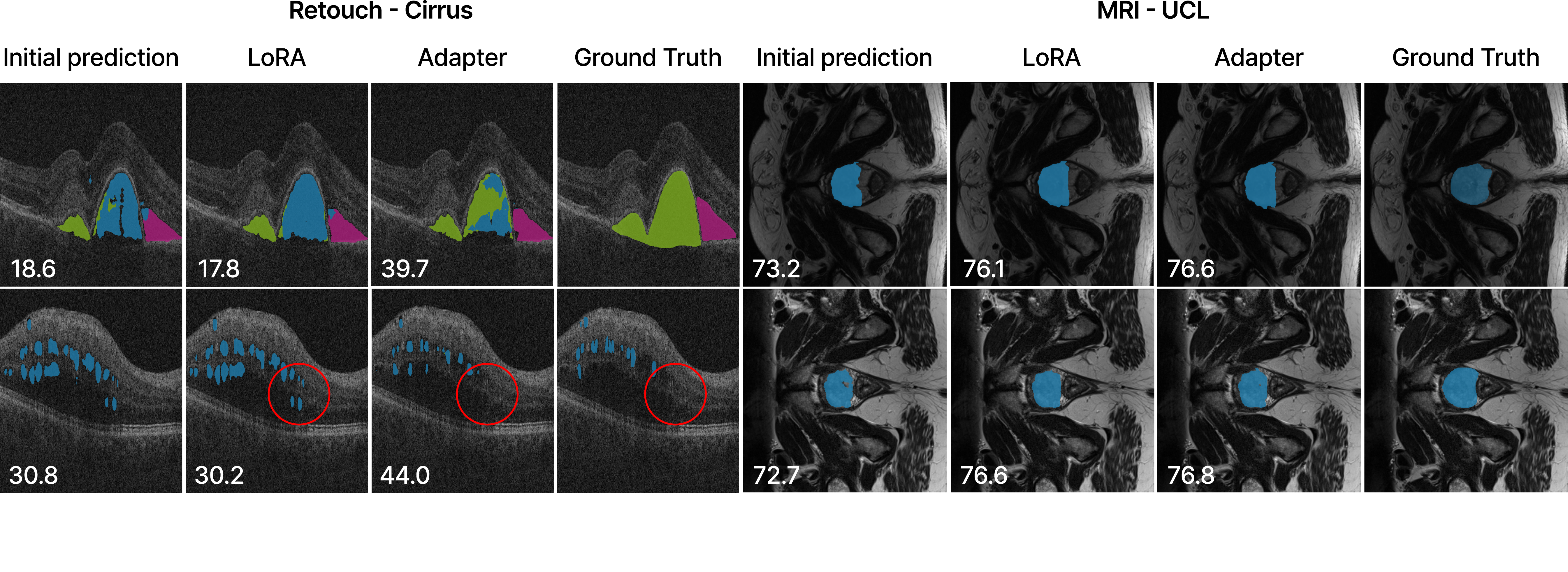}
\caption{Qualitative and quantitative results on Retouch and MRI datasets for the proposed model and LoRA. For reference, each image includes its IoU score after five TTDA iterations.
}
\label{fig:ttda}
\end{figure*}

In Fig.~\ref{fig:ttda}, we compare two runs for Retouch and MRI between our adapter and LoRA (the second best) after five iterations. 
\section{Conclusion}
\label{sec:conclusion}
In this work, we propose a SAM Decoder Adapter for semantic segmentation that introduces negligible overhead. We achieve this by using a lean approach, using the image embeddings in the decoder as queries in the attention modules of the adapter, and combining the result before the next two-way attention layer. We outperform the mask prediction quality of state-of-the-art methods and show that zero-shot generalization capabilities are improved. Furthermore, we evaluate our method on a more challenging task, Test-Time Domain Adaptation, and show its superiority against other large model adapters. With extensive ablation studies, we explain the design choices behind this simple yet powerful adapter.

\newpage

{\small
\bibliographystyle{ieee_fullname}
\bibliography{bibliography}
}

\newpage
\onecolumn
\appendix
\section{Implementation details}
\label{sec:impl}
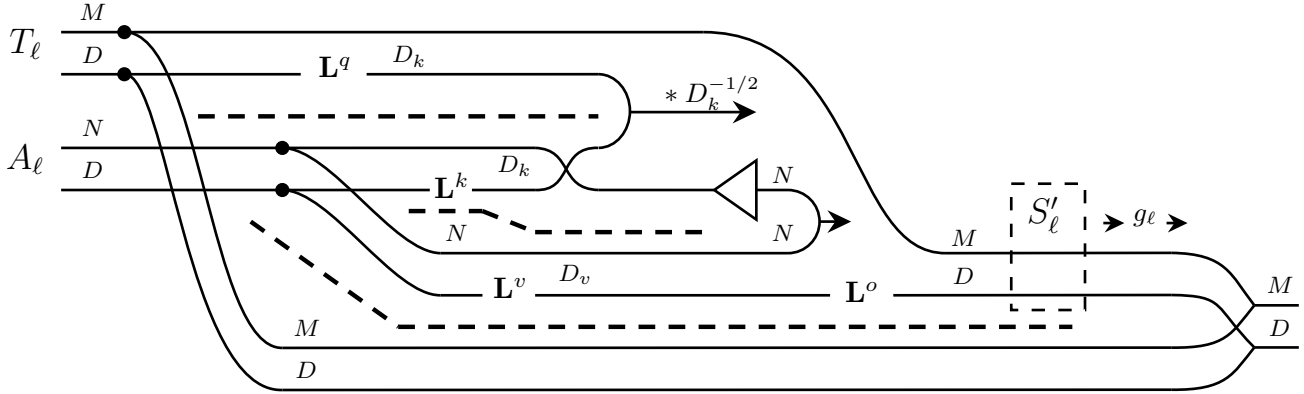
\begin{figure}[h]
    \centering
    \resizebox{\columnwidth}{!}{%
            
\tikzset{every picture/.style={line width=0.9pt}} 

\begin{tikzpicture}[x=0.75pt,y=0.75pt,yscale=-1,xscale=1]

\draw [color={rgb, 255:red, 0; green, 0; blue, 0 }  ,draw opacity=1 ]   (28.52,24.09) -- (317.42,24.09) ;
\draw [color={rgb, 255:red, 0; green, 0; blue, 0 }  ,draw opacity=1 ]   (284.27,60.08) -- (338.1,60.07) ;
\draw [shift={(341.1,60.07)}, rotate = 179.99] [fill={rgb, 255:red, 0; green, 0; blue, 0 }  ,fill opacity=1 ][line width=0.08]  [draw opacity=0] (10.72,-5.15) -- (0,0) -- (10.72,5.15) -- (7.12,0) -- cycle    ;
\draw [color={rgb, 255:red, 0; green, 0; blue, 0 }  ,draw opacity=1 ]   (28.52,43.04) -- (136.25,43.04) ;
\draw [color={rgb, 255:red, 0; green, 0; blue, 0 }  ,draw opacity=1 ]   (28.52,76.19) -- (127.98,76.19) ;
\draw [color={rgb, 255:red, 0; green, 0; blue, 0 }  ,draw opacity=1 ]   (28.52,95.13) -- (127.98,95.13) ;
\draw [color={rgb, 255:red, 0; green, 0; blue, 0 }  ,draw opacity=1 ][line width=1.5]  [dash pattern={on 5.63pt off 4.5pt}]  (90.09,61.98) -- (270.06,61.98) ;
\draw [color={rgb, 255:red, 0; green, 0; blue, 0 }  ,draw opacity=1 ]   (199.02,123.55) -- (355.31,123.55) ;
\draw  [draw opacity=0] (270.28,76.17) .. controls (278.02,76.03) and (284.27,68.67) .. (284.27,59.6) .. controls (284.27,50.44) and (277.91,43.02) .. (270.06,43.02) .. controls (270.04,43.02) and (270.01,43.02) .. (269.99,43.02) -- (270.06,59.6) -- cycle ; \draw  [color={rgb, 255:red, 0; green, 0; blue, 0 }  ,draw opacity=1 ] (270.28,76.17) .. controls (278.02,76.03) and (284.27,68.67) .. (284.27,59.6) .. controls (284.27,50.44) and (277.91,43.02) .. (270.06,43.02) .. controls (270.04,43.02) and (270.01,43.02) .. (269.99,43.02) ;  
\draw [color={rgb, 255:red, 0; green, 0; blue, 0 }  ,draw opacity=1 ]   (241.64,76.17) .. controls (258.95,76.9) and (252,94.58) .. (270.06,95.12) ;
\draw [color={rgb, 255:red, 0; green, 0; blue, 0 }  ,draw opacity=1 ]   (241.64,95.12) .. controls (258.95,95.84) and (252,75.63) .. (270.06,76.17) ;
\draw [color={rgb, 255:red, 0; green, 0; blue, 0 }  ,draw opacity=1 ]   (270.06,95.12) -- (322.16,95.13) ;
\draw  [draw opacity=0] (355.31,123.55) .. controls (363.16,123.55) and (369.52,117.19) .. (369.52,109.34) .. controls (369.52,101.5) and (363.16,95.13) .. (355.31,95.13) -- (355.31,109.34) -- cycle ; \draw  [color={rgb, 255:red, 0; green, 0; blue, 0 }  ,draw opacity=1 ] (355.31,123.55) .. controls (363.16,123.55) and (369.52,117.19) .. (369.52,109.34) .. controls (369.52,101.5) and (363.16,95.13) .. (355.31,95.13) ;  
\draw [color={rgb, 255:red, 0; green, 0; blue, 0 }  ,draw opacity=1 ]   (341.1,95.13) -- (355.31,95.13) ;
\draw   (322.16,95.13) -- (341.1,81.81) -- (341.1,108.45) -- cycle ;
\draw [color={rgb, 255:red, 0; green, 0; blue, 0 }  ,draw opacity=1 ]   (369.52,109.34) -- (380.73,109.34) ;
\draw [shift={(383.73,109.34)}, rotate = 180] [fill={rgb, 255:red, 0; green, 0; blue, 0 }  ,fill opacity=1 ][line width=0.08]  [draw opacity=0] (10.72,-5.15) -- (0,0) -- (10.72,5.15) -- (7.12,0) -- cycle    ;
\draw [color={rgb, 255:red, 0; green, 0; blue, 0 }  ,draw opacity=1 ]   (426.35,123.55) -- (527.64,123.42) ;
\draw [color={rgb, 255:red, 0; green, 0; blue, 0 }  ,draw opacity=1 ]   (317.42,24.09) .. controls (387.99,24.27) and (382.54,123.97) .. (426.35,123.55) ;
\draw [color={rgb, 255:red, 0; green, 0; blue, 0 }  ,draw opacity=1 ]   (127.98,76.19) .. controls (149.75,76.35) and (177.86,123.17) .. (199.02,123.55) ;
\draw [shift={(127.98,76.19)}, rotate = 0.42] [color={rgb, 255:red, 0; green, 0; blue, 0 }  ,draw opacity=1 ][fill={rgb, 255:red, 0; green, 0; blue, 0 }  ,fill opacity=1 ][line width=0.75]      (0, 0) circle [x radius= 2.68, y radius= 2.68]   ;
\draw [color={rgb, 255:red, 0; green, 0; blue, 0 }  ,draw opacity=1 ][line width=1.5]  [dash pattern={on 5.63pt off 4.5pt}]  (241.64,114.08) -- (322.16,114.08) ;
\draw [color={rgb, 255:red, 0; green, 0; blue, 0 }  ,draw opacity=1 ]   (127.98,95.13) .. controls (149.51,95.3) and (177.86,142.66) .. (199.02,142.5) ;
\draw [shift={(127.98,95.13)}, rotate = 0.43] [color={rgb, 255:red, 0; green, 0; blue, 0 }  ,draw opacity=1 ][fill={rgb, 255:red, 0; green, 0; blue, 0 }  ,fill opacity=1 ][line width=0.75]      (0, 0) circle [x radius= 2.68, y radius= 2.68]   ;
\draw [color={rgb, 255:red, 0; green, 0; blue, 0 }  ,draw opacity=1 ]   (165.87,43.04) -- (269.99,43.02) ;
\draw [color={rgb, 255:red, 0; green, 0; blue, 0 }  ,draw opacity=1 ]   (213.23,95.13) -- (241.64,95.13) ;
\draw [color={rgb, 255:red, 0; green, 0; blue, 0 }  ,draw opacity=1 ]   (199.02,142.5) -- (217.96,142.5) ;
\draw [color={rgb, 255:red, 0; green, 0; blue, 0 }  ,draw opacity=1 ]   (241.64,142.5) -- (374.25,142.5) ;
\draw [color={rgb, 255:red, 0; green, 0; blue, 0 }  ,draw opacity=1 ][line width=1.5]  [dash pattern={on 5.63pt off 4.5pt}]  (184.81,104.61) -- (217.96,104.61) ;
\draw [color={rgb, 255:red, 0; green, 0; blue, 0 }  ,draw opacity=1 ][line width=1.5]  [dash pattern={on 5.63pt off 4.5pt}]  (217.96,104.61) -- (241.64,114.08) ;
\draw [color={rgb, 255:red, 0; green, 0; blue, 0 }  ,draw opacity=1 ]   (402.67,142.5) -- (527.64,142.37) ;
\draw  [dash pattern={on 4.5pt off 4.5pt}] (455.71,92.29) -- (488.87,92.29) -- (488.87,149.13) -- (455.71,149.13) -- cycle ;
\draw [color={rgb, 255:red, 0; green, 0; blue, 0 }  ,draw opacity=1 ]   (56.94,43.04) .. controls (76.07,42.2) and (85.17,186.36) .. (127.98,185.12) ;
\draw [shift={(56.94,43.04)}, rotate = 357.48] [color={rgb, 255:red, 0; green, 0; blue, 0 }  ,draw opacity=1 ][fill={rgb, 255:red, 0; green, 0; blue, 0 }  ,fill opacity=1 ][line width=0.75]      (0, 0) circle [x radius= 2.68, y radius= 2.68]   ;
\draw [color={rgb, 255:red, 0; green, 0; blue, 0 }  ,draw opacity=1 ]   (56.94,24.09) .. controls (91.23,24.01) and (95.02,166.66) .. (127.98,166.18) ;
\draw [shift={(56.94,24.09)}, rotate = 359.86] [color={rgb, 255:red, 0; green, 0; blue, 0 }  ,draw opacity=1 ][fill={rgb, 255:red, 0; green, 0; blue, 0 }  ,fill opacity=1 ][line width=0.75]      (0, 0) circle [x radius= 2.68, y radius= 2.68]   ;
\draw [color={rgb, 255:red, 0; green, 0; blue, 0 }  ,draw opacity=1 ]   (127.98,166.18) -- (527.64,166.05) ;
\draw [color={rgb, 255:red, 0; green, 0; blue, 0 }  ,draw opacity=1 ]   (127.98,185.12) -- (527.64,184.99) ;
\draw [color={rgb, 255:red, 0; green, 0; blue, 0 }  ,draw opacity=1 ]   (525.53,110) -- (532,110) ;
\draw [shift={(535,110)}, rotate = 180] [fill={rgb, 255:red, 0; green, 0; blue, 0 }  ,fill opacity=1 ][line width=0.08]  [draw opacity=0] (8.93,-4.29) -- (0,0) -- (8.93,4.29) -- (5.93,0) -- cycle    ;
\draw [color={rgb, 255:red, 0; green, 0; blue, 0 }  ,draw opacity=1 ]   (497.11,110) -- (503.58,110) ;
\draw [shift={(506.58,110)}, rotate = 180] [fill={rgb, 255:red, 0; green, 0; blue, 0 }  ,fill opacity=1 ][line width=0.08]  [draw opacity=0] (8.93,-4.29) -- (0,0) -- (8.93,4.29) -- (5.93,0) -- cycle    ;
\draw [color={rgb, 255:red, 0; green, 0; blue, 0 }  ,draw opacity=1 ]   (527.64,123.42) .. controls (552.56,123.76) and (554.33,133.2) .. (565,147.02) ;
\draw [color={rgb, 255:red, 0; green, 0; blue, 0 }  ,draw opacity=1 ]   (527.64,166.05) .. controls (552.56,166.38) and (554.78,159.65) .. (565,147.02) ;
\draw [color={rgb, 255:red, 0; green, 0; blue, 0 }  ,draw opacity=1 ]   (527.64,142.37) .. controls (552.56,142.7) and (546.42,147.31) .. (565,165.97) ;
\draw [color={rgb, 255:red, 0; green, 0; blue, 0 }  ,draw opacity=1 ]   (527.64,184.99) .. controls (552.56,185.32) and (555.44,177.87) .. (565,165.97) ;
\draw    (565,147.02) -- (585,147.02) ;
\draw    (565,165.97) -- (585,165.97) ;
\draw [color={rgb, 255:red, 0; green, 0; blue, 0 }  ,draw opacity=1 ][line width=1.5]  [dash pattern={on 5.63pt off 4.5pt}]  (180.08,156.7) -- (483.18,156.7) ;
\draw [color={rgb, 255:red, 0; green, 0; blue, 0 }  ,draw opacity=1 ]   (127.98,95.13) -- (194.28,95.13) ;
\draw [color={rgb, 255:red, 0; green, 0; blue, 0 }  ,draw opacity=1 ]   (127.98,76.19) -- (241.64,76.17) ;
\draw [color={rgb, 255:red, 0; green, 0; blue, 0 }  ,draw opacity=1 ][line width=1.5]  [dash pattern={on 5.63pt off 4.5pt}]  (113.77,109.34) -- (180.08,156.7) ;

\draw (35.23,9.94) node [anchor=north west][inner sep=0.75pt]  [font=\fontsize{0.82em}{0.99em}\selectfont]  {$M$};
\draw (3.76,21.05) node [anchor=north west][inner sep=0.75pt]  [font=\large]  {$T_{\ell }$};
\draw (36.23,28.89) node [anchor=north west][inner sep=0.75pt]  [font=\fontsize{0.82em}{0.99em}\selectfont]  {$D$};
\draw (2.26,73.14) node [anchor=north west][inner sep=0.75pt]  [font=\large]  {$A_{\ell }$};
\draw (36.23,80.98) node [anchor=north west][inner sep=0.75pt]  [font=\fontsize{0.82em}{0.99em}\selectfont]  {$D$};
\draw (36.23,62.04) node [anchor=north west][inner sep=0.75pt]  [font=\fontsize{0.82em}{0.99em}\selectfont]  {$N$};
\draw (298.21,41.97) node [anchor=north west][inner sep=0.75pt]  [font=\small]  {$*\ D_{k}^{-1/2}$};
\draw (142.61,33.5) node [anchor=north west][inner sep=0.75pt]  [font=\normalsize]  {$\mathbf{L}^{q}$};
\draw (175.79,28.86) node [anchor=north west][inner sep=0.75pt]  [font=\fontsize{0.82em}{0.99em}\selectfont]  {$D_{k}$};
\draw (194.7,85.59) node [anchor=north west][inner sep=0.75pt]  [font=\normalsize]  {$\mathbf{L}^{k}$};
\draw (223.15,78.12) node [anchor=north west][inner sep=0.75pt]  [font=\fontsize{0.82em}{0.99em}\selectfont]  {$D_{k}$};
\draw (221.22,132.95) node [anchor=north west][inner sep=0.75pt]  [font=\normalsize]  {$\mathbf{L}^{v}$};
\draw (250.61,127.37) node [anchor=north west][inner sep=0.75pt]  [font=\fontsize{0.82em}{0.99em}\selectfont]  {$D_{v}$};
\draw (199.62,109.4) node [anchor=north west][inner sep=0.75pt]  [font=\fontsize{0.82em}{0.99em}\selectfont]  {$N$};
\draw (346.44,82.88) node [anchor=north west][inner sep=0.75pt]  [font=\fontsize{0.82em}{0.99em}\selectfont]  {$N$};
\draw (346.44,109.4) node [anchor=north west][inner sep=0.75pt]  [font=\fontsize{0.82em}{0.99em}\selectfont]  {$N$};
\draw (461.29,96.82) node [anchor=north west][inner sep=0.75pt]  [font=\large]  {$S'_{\ell }$};
\draw (379.41,134.85) node [anchor=north west][inner sep=0.75pt]  [font=\normalsize]  {$\mathbf{L}^{o}$};
\draw (426.9,111.29) node [anchor=north west][inner sep=0.75pt]  [font=\fontsize{0.82em}{0.99em}\selectfont]  {$M$};
\draw (427.9,128.34) node [anchor=north west][inner sep=0.75pt]  [font=\fontsize{0.82em}{0.99em}\selectfont]  {$D$};
\draw (509,102.4) node [anchor=north west][inner sep=0.75pt]  [font=\small]  {$g_{\ell }$};
\draw (569,132.4) node [anchor=north west][inner sep=0.75pt]  [font=\fontsize{0.82em}{0.99em}\selectfont]  {$M$};
\draw (570,151.82) node [anchor=north west][inner sep=0.75pt]  [font=\fontsize{0.82em}{0.99em}\selectfont]  {$D$};
\draw (131.37,152.02) node [anchor=north west][inner sep=0.75pt]  [font=\fontsize{0.82em}{0.99em}\selectfont]  {$M$};
\draw (132.37,170.97) node [anchor=north west][inner sep=0.75pt]  [font=\fontsize{0.82em}{0.99em}\selectfont]  {$D$};
\end{tikzpicture}
}
\caption{Neural circuit diagram for the proposed SAM-Decoder-Adapter
}
\label{fig:neural_diagram}
\end{figure}

Figure~\ref{fig:neural_diagram} shows the architecture of our adapter ($A_\ell$) as a neural diagram (introduced in \cite{abbott2024neural}). It is combined with the embeddings $T_\ell$ in an attention module in which $T_\ell$ act as queries and the adapter weights~$A_\ell$ act as keys and values.

\section{Further results}

Table~\ref{tab:hqseg_results} shows the IoU of all the methods on the four different subsets that compose HQSeg-44K~\cite{ke2024segment}. Encoder Adapter and Decoder Adapter represent our proposed methods with two different placements of the adapter (encoder and decoder, respectively). Due to the high number of images, we see that adaption methods with a higher number of parameters, such as LoRA, outperform smaller ones, such as HQ-SAM or our SAM Adapter. 

\begin{table}[h]
\centering
\begin{tabular}{@{}lP{2.2cm}P{2.2cm}P{2.2cm}P{2.2cm}@{}}
\toprule
 & COIFT & HRSOD & ThinObject5k & DIS5K \\ \midrule
Fine-Tuning & $82.07{\scriptscriptstyle \pm 0.58}$ & $79.44{\scriptscriptstyle \pm 0.14}$ & $79.32{\scriptscriptstyle \pm 0.33}$ & $63.33{\scriptscriptstyle \pm 0.34}$ \\
Decoder FT & $84.91{\scriptscriptstyle \pm 0.58}$ & $82.46{\scriptscriptstyle \pm 0.14}$ & $84.54{\scriptscriptstyle \pm 0.33}$ & $71.61{\scriptscriptstyle \pm 0.34}$ \\
LoRA & $86.01{\scriptscriptstyle \pm 0.19}$ & $84.50{\scriptscriptstyle \pm 0.18}$ & $87.86{\scriptscriptstyle \pm 0.26}$ & $74.22{\scriptscriptstyle \pm 0.52}$ \\
Med-SA & $85.74{\scriptscriptstyle \pm 0.15}$ & $83.94{\scriptscriptstyle \pm 0.73}$ & $89.83{\scriptscriptstyle \pm 0.31}$ & $75.69{\scriptscriptstyle \pm 0.28}$ \\
HQ-SAM & $84.17{\scriptscriptstyle \pm 0.20}$ & $81.41{\scriptscriptstyle \pm 0.26}$ & $81.41{\scriptscriptstyle \pm 0.12}$ & $69.88{\scriptscriptstyle \pm 0.12}$ \\ \midrule
Encoder Adapter & $84.82{\scriptscriptstyle \pm 0.26}$ & $82.41{\scriptscriptstyle \pm 0.30}$ & $84.61{\scriptscriptstyle \pm 0.15}$ & $71.41{\scriptscriptstyle \pm 0.42}$ \\
Decoder Adapter & $84.61{\scriptscriptstyle \pm 0.28}$ & $81.81{\scriptscriptstyle \pm 0.50}$ & $82.73{\scriptscriptstyle \pm 0.10}$ & $69.24{\scriptscriptstyle \pm 0.56}$ \\ \bottomrule
\end{tabular}
\caption{IoU of all the methods on the different subsets that compose HQSeg-44K. Variance has been obtained over four trained models on the validation set}
\label{tab:hqseg_results}
\end{table}

\begin{figure}[t]
\centering
\includegraphics[width=0.85\textwidth]{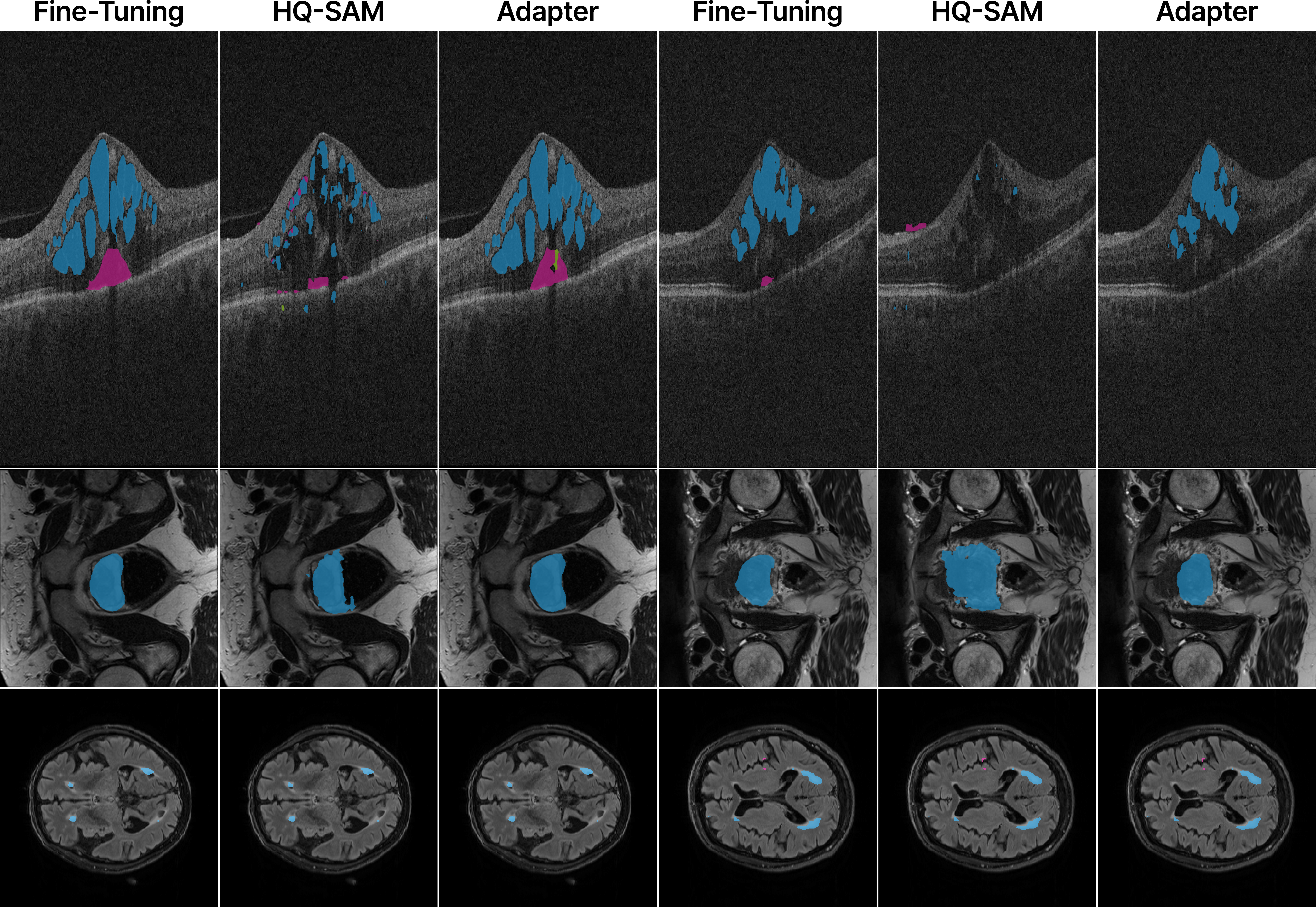}
\caption{Qualitative results on three randomly selected test samples from domain generalization subsets
}
\label{fig:generalization}
\end{figure}

Figure~\ref{fig:generalization} shows qualitative results on three test samples selected randomly from the untrained domains (Cirrus for Retouch~\cite{retouch}, UCL for MRI~\cite{liu2020shape,mri_dataset}, and NUHS Singapore for WMH~\cite{wmh}).

Tables~\ref{tab:supp_ablation_size} and~\ref{tab:supp_ablation_size_transfer} show the impact of the dimension of the adaption prompt $A_\ell$ on the performance.

\begin{table}[h]
\centering
\begin{tabular}{@{}lcccc@{}}
\toprule
 & Retouch - Spectralis & MRI - BMC & WMH - Utrecht & HQ-Seg \\ \midrule
512 & $75.4{\scriptscriptstyle \pm 0.6}$ & $86.2{\scriptscriptstyle \pm 1.5}$ & $44.2{\scriptscriptstyle \pm 0.3}$ & $79.6{\scriptscriptstyle \pm 0.4}$ \\
1024 & $75.8{\scriptscriptstyle \pm 0.8}$ & $85.6{\scriptscriptstyle \pm 1.6}$ & $40.5{\scriptscriptstyle \pm 3.8}$ & $71.6{\scriptscriptstyle \pm 2.8}$ \\
2048 & $76.3{\scriptscriptstyle \pm 0.8}$ & $85.2{\scriptscriptstyle \pm 1.7}$ & $39.0{\scriptscriptstyle \pm 4.7}$ & $70.2{\scriptscriptstyle \pm 2.3}$ \\ \bottomrule
\end{tabular}
\caption{Ablation study for the size of the adapter embeddings. IoU scores for full supervision. Variances are computed over four trained models tested on the testing set.}
\label{tab:supp_ablation_size}
\end{table}

\begin{table}[h]
\centering
\begin{tabular}{@{}lccc@{}}
\toprule
 & Retouch - Cirrus & MRI - UCL & WMH - Singapore \\ \midrule
512 & $70.2{\scriptscriptstyle \pm 3.1}$ & $80.6{\scriptscriptstyle \pm 1.0}$ & $39.6{\scriptscriptstyle \pm 0.7}$ \\
1024 & $69.4{\scriptscriptstyle \pm 1.3}$ & $79.8{\scriptscriptstyle \pm 1.3}$ & $38.7{\scriptscriptstyle \pm 3.1}$ \\
2048 & $69.2{\scriptscriptstyle \pm 1.2}$ & $80.2{\scriptscriptstyle \pm 1.1}$ & $35.5{\scriptscriptstyle \pm 3.4}$ \\ \bottomrule
\end{tabular}
\caption{Ablation study for the size of the adapter embeddings. Application of trained models to zero-shot domain generalization. The variance was obtained over four trained models tested on the testing set.}
\label{tab:supp_ablation_size_transfer}
\end{table}
\section{Statistical Analysis}

To evaluate the statistical significance of performance differences between Med-SA~\cite{wu2023medical} and SAM-DA on the fully supervised task, we conducted a paired t-test on image-wise mIoU scores obtained from both methods on the Retouch-Spectralis dataset. The test was conducted under the null hypothesis that Med-SA and SAM-DA achieve the same mean mIoU across images. A p-value of $p<0.01$ was obtained, providing sufficient evidence to reject the null hypothesis and indicating a significant performance difference between the methods.

As each image may contain a unique subset of classes, image-wise mIoU scores are not directly comparable across images, and the application of this paired t-test is not completely justified from a theoretical standpoint. However, this approach was selected as the most feasible option among available alternatives, despite its limitations. Alternative paired t-test methods were considered but ultimately dismissed as unworkable. For instance, conducting a t-test over multiple random seeds per model would have required more than 100 seeds to reach a statistical power of 0.9, which was impractical. A pixel-level paired t-test was also ruled out due to the high correlation between pixels within images, which would likely yield an artificially low and unreliable p-value.

\end{document}